\pdfoutput=1

\documentclass[11pt]{article}

\usepackage[]{ACL2023}

\usepackage{times}
\usepackage{latexsym}

\usepackage[T1]{fontenc}

\usepackage[utf8]{inputenc}

\usepackage{microtype}

\usepackage{inconsolata}

\usepackage[english]{babel}
\usepackage{framed}
\usepackage[normalem]{ulem}
\usepackage{algorithm, algorithmic}
\usepackage{amsmath, amsthm, amssymb,amsfonts}
\usepackage{enumerate}
\usepackage{natbib}
\usepackage{xcolor}
\usepackage{comment}
\usepackage{enumitem}

\usepackage{multirow} 
\usepackage{graphicx}
\usepackage{subcaption}
\usepackage{booktabs} %
\usepackage{hyperref}
\usepackage{pifont}%
\usepackage{todonotes}
\usepackage{tabularray}

\title{DEM: Distribution Edited Model for Training with \\ Mixed Data Distributions}
\author{
  Dhananjay Ram~$^\spadesuit$ \quad  Aditya Rawal~$^\spadesuit$ \quad  Momchil Hardalov~$^\clubsuit$ \\
  \textbf{Nikolaos Pappas~$^\clubsuit$ \quad  Sheng Zha~$^\spadesuit$}  \\
  $^\spadesuit$AGI Foundations, Amazon \quad $^\clubsuit$AWS AI Labs \\
  \texttt{\{radhna, adirawal, momchilh, nppappa, zhasheng\}@amazon.com} \\}

\begin{document}
\maketitle
\begin{abstract}

Training with mixed data distributions is a common and important part of creating multi-task and instruction-following models. The diversity of the data distributions and cost of joint training makes the optimization procedure extremely challenging.~Data mixing methods partially address this problem, albeit having a sub-optimal performance across data sources and require multiple expensive training runs. In this paper, we propose a simple and efficient alternative for better optimization of the data sources by combining models individually trained on each data source with the base model using basic element-wise vector operations. The resulting model, namely \emph{Distribution Edited Model (DEM)}, is $11\times$ cheaper than standard data mixing and outperforms strong baselines on a variety of benchmarks, yielding upto 6.2\% improvement on MMLU, 11.5\% on BBH, 16.1\% on DROP, 6\% on MathQA, and 9.3\% on HELM with models of size 3B to 13B. Notably, DEM does not require full re-training when modifying a single data-source, thus making it very flexible and scalable for training with diverse data sources. 
\end{abstract}

\section{Introduction}
\begin{figure}[t]
\centering  
\includegraphics[width=\columnwidth]{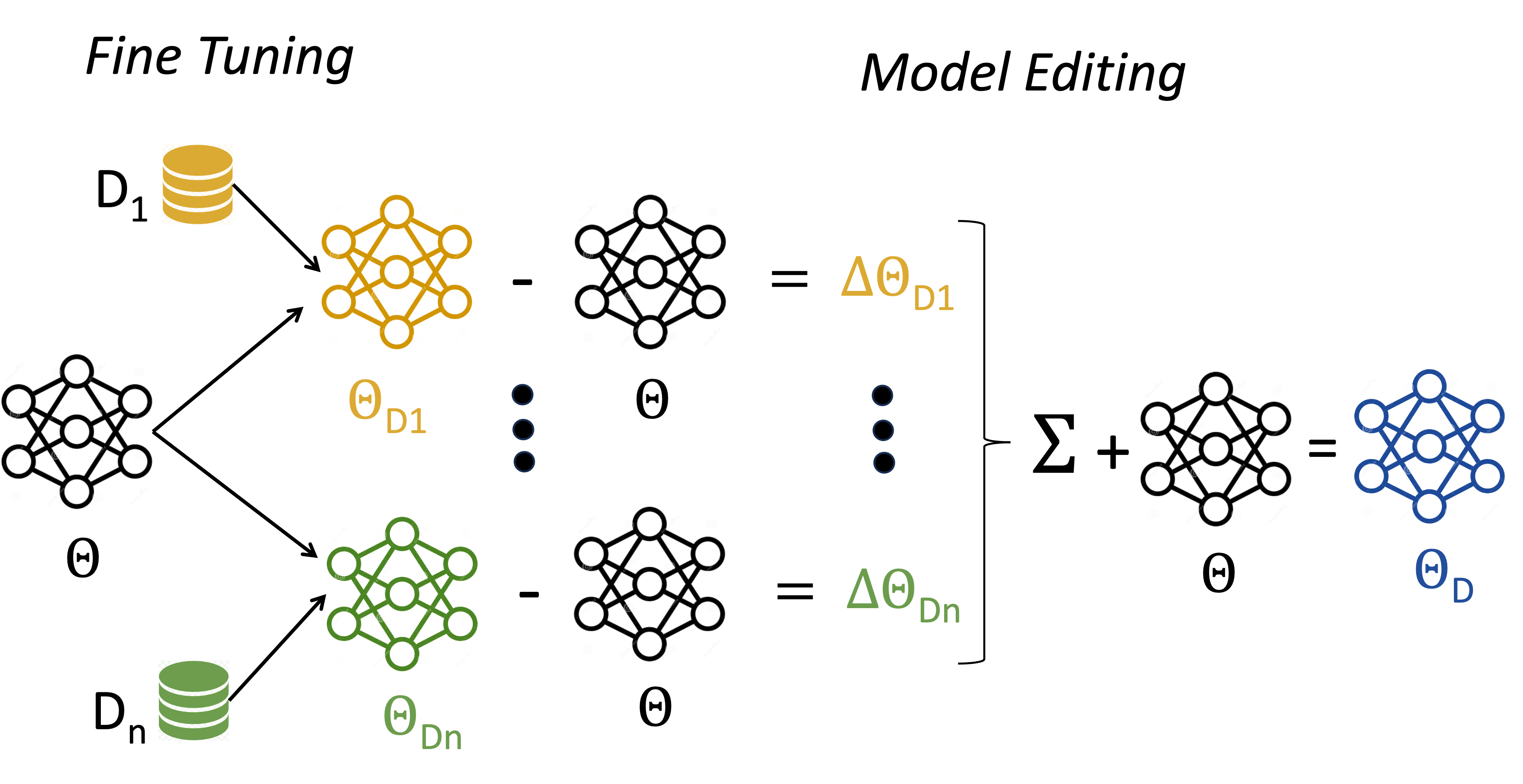}
    \caption{The \emph{Distribution Edited Model} ($\Theta_D$) results from fine-tuning a pretrained model ($\Theta$) on $n$ individual data distributions ($D_i$) and combining the resulting models with basic element-wise vector operations. Here, the combination is achieved by extracting \emph{distribution vectors} ($\Delta \Theta_{D_i}$), multiplying them by weight coefficients ($\omega_i$), and adding their weighted sum to the base model.} %
\label{fig:dem}
\end{figure} 

Large Language Models (LLM) go through an extensive pretraining on billions or trillions of tokens~\citep{brown2020language,zhang2022opt,raffel2020exploring,touvron2023llama,touvron2023llama2,openlm2023openllama}, but they typically require supervised fine-tuning on diverse instruction-following datasets for properly following human instructions~\cite{ouyang2022training, sanh2021multitask, iyer2022opt, JMLR:v25:23-0870}. Supervised training is crucial for ensuring that generated outputs meet user expectations and perform well on downstream tasks~\cite{Radford2019LanguageMA,gururangan-etal-2020-dont}.

The datasets for supervised training are often of different sizes and follow different distributions. Recent state-of-the-art fine-tuning approaches~\cite{iyer2022opt,JMLR:v25:23-0870} demonstrate that training on multiple data distributions requires careful tuning of the mixing weights for each data source to capture the combined distribution and improve downstream task performance. Tuning these weights in a data-mixing approach is a computationally expensive process. Although, there are techniques to speed-up the search~\citep{xie2023doremi, albalak2023efficient}, the process remains time-consuming. Moreover, when one or more new datasets are introduced, the weights for each dataset need to be re-tuned. This requirement makes the data-mixing approach inflexible and hard to maintain in a production environment. 

To address these challenges when fine-tuning an LLM on a set of diverse data distributions, we propose a simple and efficient approach that combines individually trained versions of the base model using element-wise vector operations. Our method focuses on the challenging setting of combining diverse data distributions that correspond to 
multiple tasks from different domains such as math, reasoning, conversations and coding.
In particular, our goal is to better capture a diverse data distributions as opposed to editing the model on a single downstream task~\cite{ilharco2022editing, schumann-etal-2024-backward}. Hence, we call resulting model \emph{Distribution Edited Model} (\emph{DEM}, shown in Figure~\ref{fig:dem}). Our experiments on a variety of downstream tasks show that \emph{DEM} is an
effective, highly capable and low cost alternative to the models trained using data mixing methods.

The primary benefit of the proposed approach is its ability to efficiently identify the optimal combination of data sources for training a model. Instead of exhaustively training and validating on all possible combinations of data sources, which can be computationally expensive, we take a more streamlined approach. 
First, we finetune the original model on each individual data source independently with early stopping to obtain the optimal model. 
Second, we extract source distribution vectors by subtracting the original model from the finetuned ones. Lastly, we create the final model by adding a weighted combination of these distribution vectors to the base model, allowing it to capture the joint distribution of different data sources in a controlled manner while enabling incremental updates with new datasets.

\noindent Our contributions can be summarized as follows:

\begin{itemize}%
\item We propose a simple and efficient approach for training models on diverse data distributions that offers a flexible way for tuning the contributions of each data source individually without the need of full data re-training (Section~\ref{sec:proposal}).
\item We show that \emph{DEM} reduces the training cost by $11\times$ 
while improving the model performance.
Compared to standard \emph{data mixing} approaches, DEM yields up to 6.2\% improvement on MMLU, 11.5\% on BBH, 16.1\% on DROP, 6\% on MathQA and 9.3\% on HELM with 3B, 7B, and 13B models.
\item We perform an exhaustive analysis of the properties of the distribution vectors and their corresponding models, finding that DEM is better aligned with the individual models than baseline while remaining close to the original model.
\end{itemize}

\section{Related Work}

\paragraph{Multi-task Fine Tuning}
Instruction-based multi-task fine-tuning of language models has been previously shown to improve both zero and few-shot performance on unseen tasks ~\citep{wei2022finetuned, sanh2021multitask}. Instruction-tuning data can be sourced from diverse task categories (such as math, reasoning, dialog etc), and the model performance is often sensitive to the data-mixing strategy. For example, both ~\citep{JMLR:v25:23-0870} and ~\citep{iyer2022opt} carefully tune the data-mixing weights for various training data sources. 

Hyperparameter tuning of data-mixing weights is a compute intensive process, and methods such as DoReMi ~\citep{xie2023doremi} and Online Data Mixing ~\citep{albalak2023efficient} have been proposed to speed-up the process for pretrained data-mixing either through a proxy-model training or through a multi-armed bandit approach respectively. \citet{renduchintala2024smart} used a submodular function to assign importance scores to tasks which are then used to determine the mixture weights. \citet{li2024many} built a framework to find multiple diverse solutions in the Pareto front of many objectives.
In this work, we propose an alternative strategy for training with multiple data sources by using vector arithmetic to combine models fine-tuned on individual datasets, rather than mixing training data in specific proportions.

\paragraph{Model Weight Interpolation}
Recently, model weight interpolation and task arithmetic techniques have been shown to improve the performance of pre-trained models on: single-task~\cite{izmailov2018averaging,matena2022merging,9878645,Wortsman_2022_CVPR} and multi-task~\cite{ilharco2022patching,ilharco2022editing,li2022branch,wortsman2022model,yadav2023tiesmerging,daheim2024model}, out-of-domain generalization~\cite{arpit2022ensemble,rame2022diverse,jin2023dataless,10.5555/3618408.3619598,cha2024swad}, and federated learning~\cite{mcmahan2017communication,Li2020On}. 

Going beyond simple weight averaging, \citep{matena2022merging} explored merging using Fisher-weighted averaging for improving single-task model performance by leveraging other auxiliary tasks. \citet{ilharco2022editing} presented a model merging technique based on producing task vectors and performing arithmetic operations, such as addition, subtraction to obtain a multitask checkpoint and `forget' unwanted behavior. 
\citet{daheim2024model} proposed a new uncertainty-based correction of the task vector coefficients to improve the performance by reducing the model mismatch.

While previous work focused on classification tasks in NLP or vision, we extend vector-arithmetic-based model editing to multi-task fine-tuning on diverse data distributions. Our results show that the proposed approach outperforms and is more efficient than data-mixed fine-tuning.

\section{Background: Data mixing} 
Let us consider a pretrained language model with parameters $\Theta$, and $D_1, D_2, ..., D_n$ denote $n$ different supervised fine tuning datasets. Each dataset can consist of a single or multiple tasks. The exact tasks may have an overlap between these datasets, however, the corresponding samples are unique to each dataset. 
Standard \emph{data mixing}~\cite{JMLR:v25:23-0870, iyer2022opt} methods create training batches by performing a weighted sampling from each of the training datasets $D_i$. The goal is to learn a joint data distribution that can span all training datasets. 

\section{Proposed Approach: Distribution Edited Model (DEM)}\label{sec:proposal}
In contrast to standard data mixing, we propose to learn each data distribution separately and combine them post training. In the following subsections, we present two variants of that lead to a \emph{Distribution Edited Model} that achieves this goal.

\subsection{Combined Distribution Vectors}
Let us assume a set of training data sources ($D_1, D_2, ..., D_n$). First, we fine tune our pretrained model ($\Theta$) on each of these $n$ datasets separately, with a different set of hyper-parameters (chosen for optimal validation loss). The corresponding fine-tuned models are noted as $\Theta_{D_1}, \Theta_{D_2}, ..., \Theta_{D_n}$. Next, we define a data distribution vector
(DV) $\Delta\Theta_{D_i}$ (corresponding to the dataset $D_i$) as the element-wise difference of parameters between the pretrained model $\Theta$ and a fine-tuned model $\Theta_{D_i}$, following a similar %
approach as presented in~\cite{ilharco2022editing}.
\begin{equation}~\label{eq:ddt}
\Delta\Theta_{D_i} = \Theta_{D_i} - \Theta,
\end{equation}
Instead of task specific model editing, as in prior work, we focus on a mixture of large number of diverse NLP downstream tasks. These different tasks are represented with their own data distribution and we investigate how to combine different data DVs that we can extract by fine tuning the pretrained model using data from different distributions. %

Lastly, we obtain a mixed data DV by computing a weighted combination of each $\Delta\Theta_{D_i}$ with corresponding weights $\omega_i \in \mathbb{R}$. Finally we add the pretrained model $\Theta$ to obtain our \emph{Distribution Edited Model (DEM)} as follows:
\begin{equation}~\label{eq:mdm}
\Theta_{D} = \Theta + \sum_{i=1}^n \omega_i \Delta\Theta_{D_i} .
\end{equation}

\subsection{Model Interpolation}\label{sec:interpolation}
Another way to combine the finetuned models ($\Theta_{D_i}$) is through model weight interpolation. In this case, we do not extract data distribution vectors ($\Delta\Theta_{D_i}$), but rather use the finetuned models directly that capture information about the data distribution. Specifically, we take a weighted average of all the fine tuned models ($\Theta_{D_i}$) where the weights, $\omega_i \in \mathbb{R}$ sum to 1. More formally,
\begin{equation}~\label{eq:mi}
\Theta_{D} = \sum_{i=1}^n \omega_i \Theta_{D_i}; \quad s.t. \quad  \sum_{i=1}^n \omega_i = 1.
\end{equation}
Note that, Eq~\ref{eq:mi} is a special case of Eq~\ref{eq:mdm} when $\sum_{i=1}^n \omega_i = 1$. \emph{DEM} using distribution vector (Eq~\ref{eq:mdm}) provides more flexibility in terms of choosing $\omega_i$ per data source which can yield further performance improvement (Section~\ref{sec:performance}). %

\subsection{Computational Cost} \label{sec:cost}
To better understand the advantages of \emph{DEM} over the \emph{data mixing} approach we derive a formula to measure the cost for each method.
Let us assume we have $n$ different data sources and $m$ number of weights per data source. The hyperparameter search space for both the approaches will have a total of $m^n$ weight combinations. Intuitively, searching for \emph{data mixing} weights is comparatively more expensive than \emph{DEM} since full data re-training is required for validating each weight combination. On the other hand, \emph{DEM} requires finding the optimal weights after individual training using only validation for each weight combination. 

To formalize this, assume $T$ and $V$ as the average number training and validation steps respectively. The computational complexity for the weighted \emph{data mixing} will be $O(m^n (T + V))$ and for the proposed \emph{DEM} approach will be $O (n (T + V) + (m^n V))$. We can clearly see that $O(m^n (T + V)) \geq O (n (T + V) + (m^n V))$, and with \emph{DEM} we reduce the number of training run by a factor of $m^n/n$.  %

Additionally, we can compare the exact training and validation cost of our proposed \emph{DEM} approach with the baseline. Assuming $k$ steps of training or validation and each step takes $t$ seconds, we can define the cost ($c$) in gpu-hours as follows:
\begin{equation}\label{eq:train-cost}
c = (k * t * g)/3600 
\end{equation}
where $g$ is the total number of GPUs used. The exact cost for training ($c_{train}$) and validation ($c_{val}$) depends on the corresponding values of $k$, $t$ and $g$ and generally $c_{train} \gg c_{val}$.

\section{Experimental Setup}

\subsection{Dataset} \label{sec:dataset}

Here, we list the fine-tuning datasets, we use to enhance instruction following capability of our base pre-trained LLM. Previous work has shown that they improve the instruction following capabilities of the models~\citep{JMLR:v25:23-0870,iyer2022opt,gupta2022instructdial,amini-etal-2019-mathqa,sanh2021multitask}. 
\begin{itemize}[leftmargin=*] %
\item {\bf Chain of Thoughts (CoT) \cite{wei2022chain}}: The CoT mixture~\cite{JMLR:v25:23-0870} consists of nine datasets with manually written CoT annotations. Each task in these nine datasets have ten manually composed instruction templates, and the span arithmetic reasoning, multi-hop reasoning, and natural language inference. 
\item {\bf Math QA \cite{amini-etal-2019-mathqa}}: This dataset consists of 37K math-based multiple-choice word problems. The problem set includes geometry, counting, probability, physics, gain-loss and other general math topics. 
\item {\bf Public Pool of Prompts (P3) \cite{sanh2021multitask}}:
P3 (Public Pool of Prompts) is a collection of prompted English datasets for a diverse set of NLP tasks, where each sample consists of a prompted input and a target text. Prompts can be considered as a functions that map an example from a dataset to a natural language input and target output. Promptsource~\cite{bach2022promptsource} is used to interactively create prompts and gather prompt-specific metadata like evaluation metrics. As of writing of this paper, over 2,000 prompts from 270+ datasets are publicly available on Promptsource.

\item {\bf Instruct Dial (InstDial) \cite{gupta2022instructdial}}: 
This is an instruction tuning dataset designed for dialogues. It consists of 48 different dialogue tasks from 59 open dialogue datasets which is unified in text-to-text format suitable for decoder LLMs. It has been shown improve model performance in unseen datasets, specially for dialogue related tasks.

\item {\bf Super Natural Instructions (SNI) \cite{wang2022super}}:
This dataset consists of 1,616 diverse NLP tasks in text-to-text format with instructions written by experts. It covers 76 distinct task types, including but not limited to text composition, infilling, extraction, classification, sequence tagging and paraphrasing. 
\end{itemize}

\subsection{Model Architecture}
We use OpenLLaMA~\cite{openlm2023openllama} as our base LLM, which is trained on 1T tokens from the RedPajama Dataset~\cite{together2023redpajama}. It follows the same architecture as the LLaMA model~\cite{touvron2023llama} -- a decoder-only LLM with rotary positional embedding, SwiGLU activations and RMS Norm for pre-normalization. In our experiments, we cover three different sized models: 3B, 7B and 13B (see Table~\ref{tab:arch}). We carry all ablations with the 7B model, while the 3B and 13B models are used to show generalization of the proposed approach to other sizes. The experimental results show that the properties of \emph{DEM} are present across different model sizes.
\begin{table}[t]
\centering
\resizebox{\columnwidth}{!}{%
\small
\setlength{\tabcolsep}{3pt}
\begin{tabular}{l|cccc} 
\toprule
{\bf \# Params} & {\bf Context} & {\bf Dims} & {\bf \# Heads} & {\bf \# Layers}  \\ 
\midrule 
3B & 2048 & 3200 & 32 & 26 \\
7B & 2048 & 4096 & 32 & 40 \\
13B & 2048 & 5120 & 40 & 40 \\
\bottomrule
\end{tabular}
}
\caption{Characteristics of different OpenLLaMA model sizes used in our experiments.}
\label{tab:arch}
\end{table}

\subsection{Training}
We fine-tune the OpenLLaMA model on all instruction following datasets (Section~\ref{sec:dataset}), both separately and jointly. 
We use AdamW optimizer with $\beta_1=0.9$, $\beta_2=0.95$, weight decay of 0.05, gradient clipping of 1 and a constant learning rate of 2e-5 with a 2000 step warmup. We also adjust batch size for different datasets based on the validation loss (see Appendix~\ref{sec:apx-hyp} for details).
We use a greedy sample packing approach to fit multiple training samples into a single batch sample efficiently, padding to the max sequence length without overflowing into the next sample of a batch. 
To select the optimal mixing weights for \emph{DEM} (Eq~\ref{eq:mdm}), we perform a grid search over $\omega_i$ values. For each coefficient combination we evaluate the validation losses on the five datasets (Section~\ref{sec:dataset}), and select the model that minimizes their average (see Section~\ref{sec:dem-weight} for details). We use an equal weight of $\omega_i = 0.25$ for all datasets in our experiments.
 
\subsection{Evaluation Framework}
We evaluate the instruction following capability of the models using three publicly available benchmarks, namely InstructEval~\citep{chia-etal-2024-instructeval}, LM-evaluation-harness~\citep{eval-harness} and HELM~\citep{liang2023holistic}. To have a holistic evaluation, we choose a diverse set of held-out datasets: (\emph{i})~from InstructEval -- MMLU, Big-Bench Hard and DROP, (\emph{ii})~from LM-evaluation-harness -- MathQA, and (\emph{iii})~from HELM -- twenty sets from six diverse task-groups -- Classification, ClosedbookQA, OpenbookQA, Math, Reasoning and Conversational (see Table~\ref{tab:helm}). We perform 5-shot evaluation on MMLU and HELM, and 3-shot evaluation on BBH and DROP, inline with the standardized setup and previous work.

\subsection{Baseline Models} \label{sec:baseline}
The pre-trained OpenLLaMA serves as the non instruction-tuned baseline for evaluation. Our primary instruction-tuned baseline is \emph{data mixing} -- the model is fine-tuned using a weighted mixture of 5 diverse datasets as described in~\ref{sec:dataset} following~\cite{JMLR:v25:23-0870, iyer2022opt} which has been shown to produce SOTA performance with large scale diverse datasets. This model requires finding the optimal weights corresponding to each training dataset such that the validation loss reaches optimal value for each dataset at similar number of training steps.  
We experimented with several combinations 
of weights and chose the one that leads  to the smallest validation loss 
(see Appendix~\ref{sec:apx-data-mix-weights} for details). 
Additionally, we create a simpler baseline where we concatenate all 5 training datasets and the samples are shuffled randomly during training. This technique is more cost-effective than the standard data mixing approach, as it does not require any weight optimization.   

\section{Experimental Analysis}

\subsection{Downstream Task Performance}~\label{sec:performance}
\begin{table}[t]
\resizebox{\columnwidth}{!}{%
\setlength{\tabcolsep}{3pt}
\begin{tabular}{l|cccc} 
\toprule
{\bf Models} & {\bf MMLU} & {\bf BBH} & {\bf DROP} & {\bf MathQA} \\ 
\midrule 
Open LLaMA & 40.31 & 32.84 & 24.38 & 27.71 \\
LLaMA~{\small \cite{touvron2023llama}} & 35.10 & 30.30 & - & -\\
LLaMA2~{\small \cite{touvron2023llama2}} & 45.30 & 32.60 & - & -\\
\midrule
FlanPaLM~{\small (8B) \cite{JMLR:v25:23-0870}} & 49.3 & 36.4 & - & - \\
OPT-IML~{\small (30B) \cite{iyer2022opt}} & 43.2 & 30.9 & - & - \\
OPT-IML~{\small (175B) \cite{iyer2022opt}} & 47.1 & 35.7 & - & - \\
\midrule
CoT & 41.67 & 33.98 & 24.20 & 29.31 \\
Math QA & 39.71 & 32.70 & 24.31 & 25.03 \\
P3 & 35.69 & 14.00 & 23.29 & 27.14 \\
InstDial & 39.31 & 23.09 & 21.81 & 26.60 \\
SNI & 46.55 & 35.88 & 34.53 & 28.31 \\
\midrule
Data Mixing & 47.77 & 36.38 & 32.71 & 30.35 \\
Concatenated Datasets & 43.43 & 21.34 & 23.21 & 27.71 \\
{\it DEM} - {\small Interpolation (Ours)} & 50.14 & 40.11 & 36.31 & 31.22 \\
{\it DEM} - {\small Distribution Vector (Ours)} & {\bf 50.74} & {\bf 40.56} & {\bf 37.96} & {\bf 32.16}\\
\bottomrule
\end{tabular}
}
\caption{Downstream task performance of models trained on different instruction following datasets (Section~\ref{sec:dataset}). We compare it with different pretrained and fine-tuned baselines (Section~\ref{sec:baseline}) and our proposed approach in Section~\ref{sec:proposal}. The models are of size 7B, unless specified. The performance numbers for models with citation are taken from the corresponding paper, rest are evaluated using InstructEval and LM-evaluation-harness.}
\label{tab:performance}
\end{table}

In this section, we first use the Instruct-Eval framework to evaluate the performance of both the pre-trained and fine-tuned models. The performance on MMLU, BBH and DROP is shown in Table~\ref{tab:performance}.\footnote{The low performance on BBH after training on P3 is inline with previous findings~\cite{iyer2022opt}. The T0pp 11B's~\cite{sanh2021multitask} accuracy is 13.0, after being fine-tuned only on the P3 dataset.} In addition to the pretrained OpenLLaMA model, we show the performance of LLaMA~\cite{touvron2023llama} and LLaMA2~\cite{touvron2023llama2} of same size as a reference. We also include three other supervised fine tuned models of larger sizes, namely FlanPaLM (8B) and OPT-IML (30, 175B). %

We present the performance of fine tuned models on each dataset separately and observe that the performance degrades compared to Open-LLaMA model for P3, InstDial and MathQA. On the other hand, we observe significant improvement with CoT and SNI datasets in all four task families. We compare these models with {\it data mixing} baseline and note that it performs significantly better than the pretrained OpenLLaMA, while the improvement compared to the best single dataset fine-tuned model (i.e. SNI) is much smaller, even worse for DROP. The {\it concatenated datasets} baseline performs significantly worse than {\it data mixing} method, only improving for MMLU compared to OpenLLaMA and significantly worse than SNI fine-tuned model. This highlights the importance of choosing the optimal weights for {\it data mixing} and training a strong baseline. 

\begin{table*}[t]
\centering
\resizebox{0.9\textwidth}{!}{%
\setlength{\tabcolsep}{3pt}
\begin{tabular}{l|cccccc} 
\toprule
 \textbf{Models} & \textbf{Classification} & \textbf{Closedbook QA} & \textbf{OpenBook QA} & \textbf{Math} & \textbf{Reasoning}  & \textbf{Conversational} \\ 
\midrule 
OpenLLaMA & 49.68 & 23.21 & 48.55 & {\bf 10.45} & 50.40 & 33.03 \\
\midrule
Data Mixing & 56.52 & {\bf 28.71} & 44.36 & 5.15 & 51.13 & 34.51 \\
{\it DEM} (Ours) & {\bf 56.94} & 28.24 & {\bf 54.34} & 7.78 & {\bf 53.31} & {\bf 40.22} \\
\bottomrule
\end{tabular}
}
\caption{Summary results of the HELM evaluations on held-out scenarios, grouped by task-category for the 7B model. \emph{DEM} outperforms \emph{data mixing} approach in five out of six HELM task clusters.} 
\label{tab:helm}
\end{table*}

Next, we combine the models fine tuned on single datasets with distribution vector and interpolation method using Eq~\ref{eq:mdm} and~\ref{eq:mi} respectively. The corresponding results are shown in Table~\ref{tab:performance}. We observe that both approaches perform significantly better than the best {\it data mixing} model for all 3 scenarios showing their effectiveness (see Appendix~\ref{sec:apx-perf} for MMLU performance per domain). We also compare \emph{DEM} with larger fine-tuned models (FlanPaLM (8B), OPT-IML (30B, 175B) and observe that \emph{DEM} performs better although these models were trained on a larger mix of tasks and datasets compared to our model. Additionally, \emph{DEM - Distribution Vector} performs better than \emph{DEM - Interpolation} due to more flexible choice of $\omega_i$ (Section~\ref{sec:interpolation}) and we use it in the rest of the paper.

We further expand our evaluation setting to include HELM scenarios. Here, we compare the performance of the pretrained model with {\it DEM} and {\it data mixing} model on multiple HELM held-out task clusters (see Table~\ref{tab:helm}). {\it DEM} outperforms the {\it data mixing} approach in five out of six HELM task clusters. Surprisingly, for Math task category, the fine-tuned model performance degrades as compared to the pretrained model. Closer inspection reveals that this degradation is partly due to the fact that the instruction-tuned model does not output the answer in the correct format (as expected by HELM evaluation metric). The detailed HELM evaluation results (including results on `seen' tasks) are reported in the Appendix~\ref{sec:apx-perf} (Table~\ref{tab:helm_detailed}).

\subsection{Effect of Model Size Scaling}\label{sec:perf-scale}
\begin{table}[t]
\centering
\resizebox{\columnwidth}{!}{%
\setlength{\tabcolsep}{3pt}
\begin{tabular}{c|l|cccc} 
\toprule
{\bf \# Params} & {\bf Models} & {\bf MMLU} & {\bf BBH} & {\bf DROP} & {\bf MathQA} \\ 
\midrule 
\multirow{2}{*}{3B} & Data Mixing & 41.08 & 31.36 & 25.98 & 28.54\\
 & {\it DEM} & {\bf 43.67} & {\bf 34.14} & {\bf 28.89} & {\bf 29.78} \\ \cmidrule(lr){1-6}
\multirow{2}{*}{7B} & Data Mixing & 47.77 & 36.38 & 32.71 & 30.35 \\
 & {\it DEM} & {\bf 50.74} & {\bf 40.56} & {\bf 37.96} & {\bf 32.16} \\ \cmidrule(lr){1-6}
\multirow{2}{*}{13B} & Data Mixing & 52.7 & 40.48 & 43.15 & 30.72 \\
 & {\it DEM} & {\bf 54.53} & {\bf 42.65} & {\bf 46.59} & {\bf 33.13} \\ 
\bottomrule
\end{tabular}
}
\caption{Effect of model size on the performance of the proposed approach. We observe performance improvement using \emph{DEM} for both smaller (3B) and larger (13B) models compared to \emph{Data Mixing} baseline.}
\label{tab:perf-scale}
\end{table}
We evaluate the performance of the proposed \emph{DEM} approach with increasing model sizes using Open-LLaMA 3B, 7B, and 13B models, quantifying the impact with both smaller and larger models. We trained the baseline \emph{Data Mixing} model using the method discussed in Section~\ref{sec:baseline}. On the other hand, we fine-tuned the models on each dataset separately and combined them using Eq~\ref{eq:mdm}, similar to the 7B model as discussed in Section~\ref{sec:performance}. We use the same model mixing weight of $\omega_i = 0.25$ (optimized for 7B model) for models of all sizes and present the results in Table~\ref{tab:perf-scale}. We observe that the model performance increases as we scale up the model size from 3B to 13B for both \emph{Data Mixing} and \emph{DEM}. Additionally, \emph{DEM} yields performance improvement for each model size, showing the effectiveness and generalizability of the proposed approach with model size.

\subsection{Impact of Different Training Datasets}\label{sec:perf-data}
In this section, we analyze the impact of each training dataset included in the mixture on the downstream task performance. For this, we progressively add the data distribution vector corresponding to each dataset to the base model (following Eq.~\ref{eq:mdm}) and evaluate the resulting model. We use $\omega_i=0.25$ for all datasets to keep the setup simple. The performance of the resulting models are presented in Table~\ref{tab:data-analysis}. We observe that these data sources yield different levels of performance gains, as expected. This can be due to the various levels of mismatch between the train and test distribution. We observe that combining the pretrained model with single-task distribution vectors (e.g Math QA) or smaller mix of tasks (e.g., CoT) leads to smaller improvement whereas large scale multi-task distribution vectors (e.g., P3 and SNI) yields a much larger performance gain, in comparison. It can also be due to the large diversity of tasks and samples in P3 and SNI. InstructDial is an exception, which can be due to the conversational nature of this dataset, making it very different from the evaluation tasks.

\begin{table}[t]
\centering
\resizebox{\columnwidth}{!}{%
\small
\setlength{\tabcolsep}{3pt}
\begin{tabular}{l|cccc} 
\toprule
{\bf Training Dataset} & {\bf MMLU} & {\bf BBH} & {\bf DROP} & {\bf MathQA} \\ 
\midrule 
Open LLaMA & 40.31 & 32.84 & 24.38 & 27.71 \\
+ CoT & 41.30 & 33.68 & 25.46 & 28.44 \\
+ MathQA & 41.67 & 33.73 & 26.05 & 28.68 \\
+ P3 & 47.12 & 36.58 & 30.82 & 30.35 \\
+ InstDial & 47.44 & 38.20 & 31.15 & 30.65 \\
+ SNI & {\bf 50.74} & {\bf 40.56} & {\bf 37.96} & {\bf 32.16} \\
\bottomrule
\end{tabular}
}
\caption{Effect of progressively adding distribution vectors (Eq~\ref{eq:ddt}) from different data sources to the pretrained model using {\it DEM} (Eq~\ref{eq:mdm}). The performance increases as we add more data sources.}
\label{tab:data-analysis}
\end{table}

\begin{table}[!t]
    \centering
    \resizebox{0.6\columnwidth}{!}{%
    \begin{tabular}{l|c}
    \toprule
     \textbf{OpenLLaMA} vs. & \textbf{Euclidean} \\
    \midrule
    P3 & 35.1 \\
    InstDial & 85.1 \\
    SNI & 34.1 \\
    CoT  & 3.2 \\
    MathQA & 4.1 \\
    \midrule
    Data Mixing & 74.6 \\
    {\it DEM} & 20.8 \\
    \bottomrule
    \end{tabular}
    }
    \caption{Euclidean distance between the base model (OpenLLaMA) and the fine-tuned models.}
    \label{tab:models:euclidean}
\end{table}

\subsection{Properties of Distribution Vectors} %
\label{sec:properties}
To better understand the behavior of {\it DEM}, we examine the characteristics of the fine-tuned models and their corresponding distribution vectors, as defined in  (Eq~\ref{eq:ddt}). We evaluate the similarity between models by calculating the Euclidean distance and the cosine similarity after converting their weights into a single flattened vector representation.

\noindent \textbf{Individual model distance from base}. In Table~\ref{tab:models:euclidean}, we show the Euclidean distance from the base model to each fine-tuned model. Datasets with more examples (P3, Instruct Dial, and SNI) lead to models that are further away from the base. The largest change is caused by Instruct Dial (x3 compared to the second largest), since it introduces a very specific domain (i.e., conversations), and requires higher adaptation of the model. In contrast, smaller datasets (CoT, and Math QA) only contribute small changes (3-4 points). As expected, the distribution edited model (\emph{{\it DEM}}) is closer to the base model than the models trained on the largest datasets. This is because DEM is derived from a weighted average of the individual vectors. Finally, we observe that the \textit{Data Mixing} model has significantly higher euclidean distance (x3) from the base model compared to {\it DEM}, indicating that the \textit{Data Mixing} approach introduces a larger change.

\begin{table}[t]
    \centering
    \resizebox{\columnwidth}{!}{%
    \setlength{\tabcolsep}{3pt}
    \begin{tabular}{l|ccccc}
    \toprule
    $\downarrow$ \textbf{Dist. Vector} $\rightarrow$ & \textbf{P3} & \textbf{InstDial} & \textbf{SNI} & \textbf{CoT} & \textbf{MathQA} \\
    \midrule
    InstDial & 0.07 & - &  & &\\
    SNI & 0.09 & 0.08 & - &  &\\
    CoT & 0.02 & 0.01 & 0.02 & - & \\
    MathQA & 0.01 & 0.01 & 0.01 & 1.0 & -  \\
    \midrule
    Data Mixing & 0.27 & 0.29 & 0.19 & 0.10 & 0.10 \\ 
    {\it DEM} & 0.44 & 0.87 & 0.43 & 0.06 & 0.06 \\
    \bottomrule
    \end{tabular}
    }
    \caption{Cosine similarity between distribution vectors.}
    \label{tab:vectors:cos}
\end{table}

\begin{figure}[!t]
    \centering
    \includegraphics[width=1.05\columnwidth]{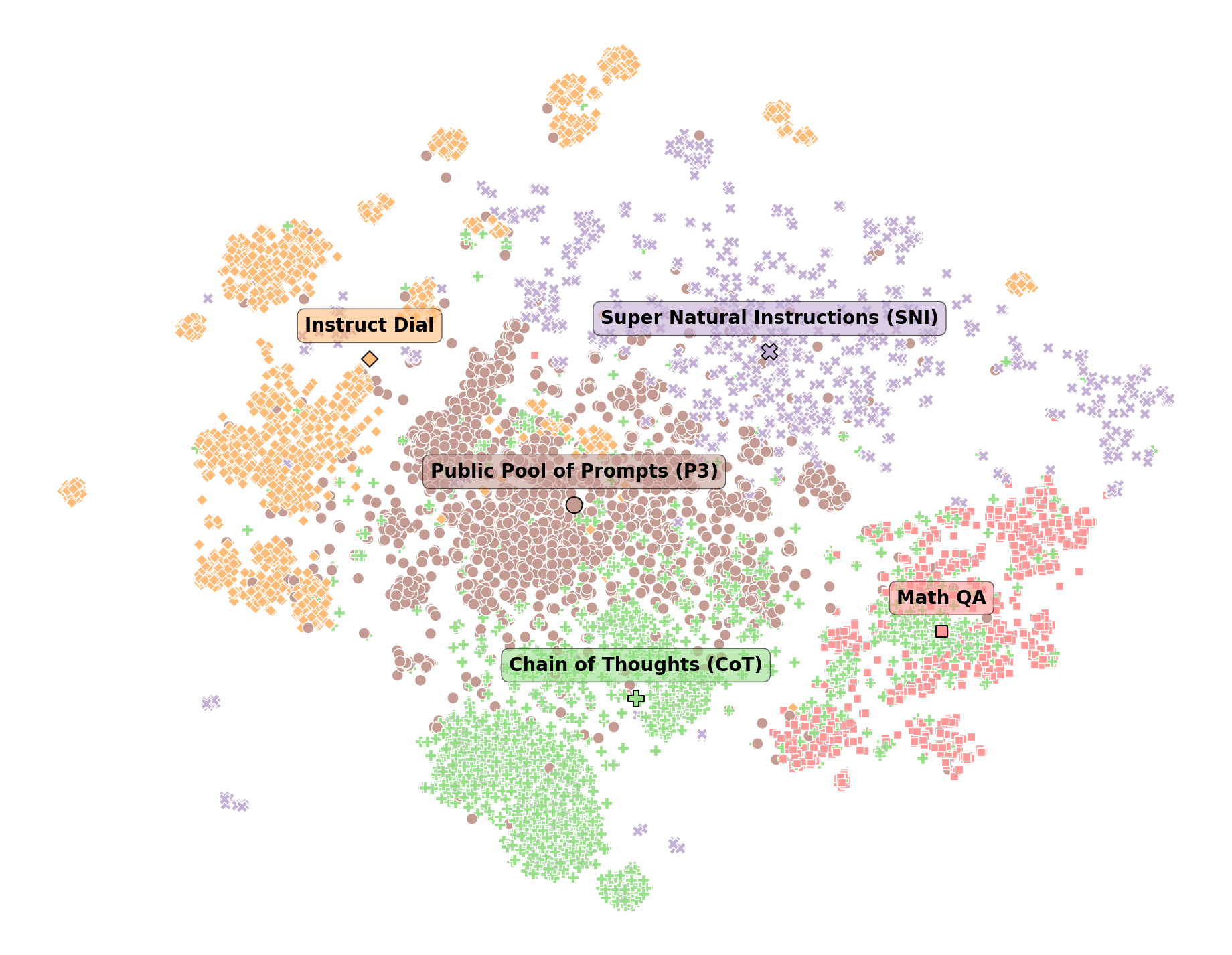}
    \caption{tSNE representation of the fine-tuning datasets. The centroids of the datasets are marked as larger points with captions. }
    \label{fig:tsne:datasets}
\end{figure}

\vspace{2mm}
\noindent \textbf{Pairwise similarity between distribution vectors}. Next, in Table~\ref{tab:vectors:cos} we compare the pairwise cosine similarity between the DVs from the fine-tuned models. We show that most of the individual DVs are almost orthogonal, except CoT and Math QA. This suggests that fine-tuning on these datasets does not lead to interference and introduces different abilities into the model.%

To understand this, we sample \emph{2,000} points from each dataset and plot their embedding representations into a common space using tSNE~(\citet{JMLR:v9:vandermaaten08a}, see Figure~\ref{fig:tsne:datasets}).\footnote{We encode all texts after formatting them into their corresponding prompt using \href{https://huggingface.co/sentence-transformers/all-MiniLM-L12-v2}{sentence-transformers/all-MiniLM-L12-v2}.} We observe a large number of CoT prompts that are close to the centroid of the MathQA dataset, which may explain the high similarity between their DVs. Note that CoT also has a small overlap with P3 but further away from their centroids, making the two DVs almost orthogonal. All other datasets have a minimal overlap between each other, and form independent clusters. We also study the relation between the combined DV and individual DVs (last row in Table~\ref{tab:vectors:cos}). We observe that \textit{DEM} is oriented towards the DVs from the models with the highest euclidean distance from the pretrained model. %

Finally, we compare how the {\it data mixing} model %
is oriented relative to the individual DVs. The cosine similarity with all DVs is less than 0.3, however, the model is oriented towards the DVs of bigger and more diverse datasets (InstDail, P3, SNI). The similarity with the CoT and MathQA is a bit higher, but it still remains within 0.1. %

\begin{figure}[t]
    \centering  
    \includegraphics[width=0.95\columnwidth]{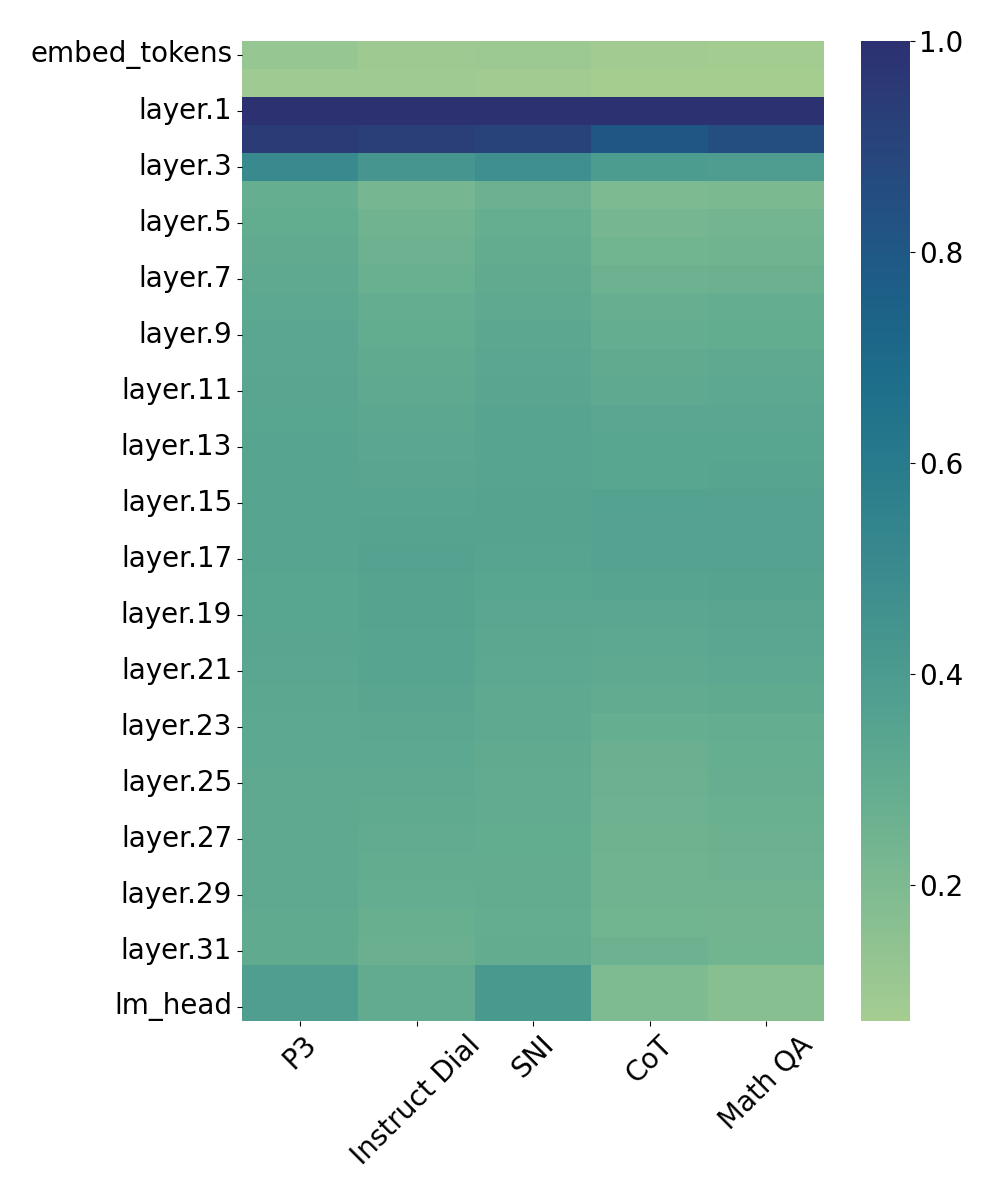}
    \caption{Layer-wise Euclidean distance, comparison between the base OpenLLaMA model, and the tuned models. Darker colors mean higher absolute difference. The euclidean distance values are normalized per-model by the highest layer-distance for that model. The plots are invariant to the scale of the weight change.}
    \label{fig:layer:euclidiean}
\end{figure}

\vspace{2mm}
\noindent \textbf{Layer-wise distance of individual models from base.} Finally, to fully understand the changes in the models and why {\it DEM} is an effective strategy for data distribution mixing, we zoom in even further into the layer-wise euclidean distance (Figure~\ref{fig:layer:euclidiean}) between the individual task vectors and the base model (OpenLLaMA 7B). From Figure~\ref{fig:layer:euclidiean},  it is evident that the changes in the tuned models occur mostly in the first three layers.
The embedding layers remain relatively stable across different domains and dataset sizes, indicating that the fundamental properties are preserved. New knowledge is primarily acquired by the 2-3 layers, which  contributes to the success of the proposed approach. Furthermore, this study suggests that when combining weights, it is not necessary to take into account all the weights involved. Instead, it is possible to safely remove or prune certain weights in the combination without significantly impacting the outcome, as also shown by \citet{yadav2023tiesmerging}.

\subsection{Compute Cost Comparison} \label{sec:cost-comp}
We use Eq.~\ref{eq:train-cost} to compare the real compute cost of the proposed {\it DEM} approach with the baseline {\it data mixing} method for 7B model on Nvidia A100 machines (with 8 gpus per node). Note that, this cost is specific to our setup and it can change depending on the model size, training parallelization scheme and other factors. In Table~\ref{tab:train-cost}, we present the gpu-hours used by different training runs, as well as the validation runs needed for finding optimal model mixing weight $\omega_i$ in Eq~\ref{eq:mdm}. In each case, we did early stopping to obtain the best validation loss, which results in varying number of training steps (`\# steps' in Table~\ref{tab:train-cost}).
As discussed in Appendix~\ref{sec:apx-data-mix-weights}, we use a combination of 10 weights to get the best model for {\it DEM}, which costs 23 gpu-hours. The total cost (training+validation) for {\it DEM} is 966 gpu-hours. 

For the baseline {\it data mixing}, we trained 50 models with different weight combination (the exact weight selection process is described in Appendix~\ref{sec:apx-data-mix-weights}). Each run costs an average of 233 gpu-hours, resulting in a total cost of 11650 gpu-hours. This is more than 11 times the total cost of  {\it DEM}.

\begin{table}[t!]
\resizebox{\columnwidth}{!}{%
\setlength{\tabcolsep}{3pt}
\begin{tabular}{l|ccc|c} 
\toprule
{\bf Train/Val Runs} & {\bf time / step} & {\bf \# steps} & {\bf \# gpus} & {\bf Cost}  \\ 
\midrule 
{\it DEM} &  &  &  &  \\
- CoT & 6.5 & 550 & 8 & 8 \\
- Math QA & 6.5 & 600 & 8 & 8.7 \\
- P3 & 4.8 & 6000 & 32 & 256 \\
- InstDial & 5.2 & 23000 & 16 & 530 \\
- SNI & 5.24 & 6000 & 16 & 140 \\

- Validation (10x) & 2.1 & 500 & 8 & 23 \\
Total &  &  &  & {\bf 966} \\
\midrule

Data-Mixing (50x) & 5.24 & 15000 & 16 & {\bf 11650} \\
\bottomrule
\end{tabular}
}
\caption{Training cost (in gpu-hours) of 7B model on different instruction following datasets computed using Eq~\ref{eq:train-cost}. Note that the number of steps is not equal to the number of examples.} %
\label{tab:train-cost}
\end{table}

\section{Conclusions and Future Work}
We proposed a simple and efficient approach for training on diverse data distributions that trains checkpoints individually on each data source and then combines them with basic element-wise vector operations. \emph{DEM} siginficantly outperforms the standard weighted data mixing in terms of downstream performance and overall compute cost. Our experiments demonstrate that \emph{DEM} works with both single-task  (e.g. Math QA) and multi-task data distributions (e.g. SNI, P3), and that they can be incrementally added to the pretrained model, resulting in improved downstream performance. We further performed extensive analysis to better understand the properties of the learned distribution vectors, finding that \emph{DEM} is better aligned with the individual models than baseline while remaining close to the original model.  

In future, it is important to evaluate the proposed approach using other model architectures e.g. encoder-decoder or mixture of experts model to better understand its effectiveness with other model designs. Additionally, \emph{DEM} can be further improved by using more sophisticated methods for combining the individual checkpoints that can reduce the negative effects of interfering data distributions.

\section*{Acknowledgments}
We thank the anonymous reviewers for their helpful questions and comments, which have helped us improve the quality of the paper. We also want to thank Yang Li for their help in setting up HELM evaluation framework, and Thomas M\"{u}ller and  Llu\'{i}s M\`arquez for helpful discussions.

\section*{Limitations}
While this paper proposes an efficient and effective alternative to data mixing for training multi-task and instruction-following models, it is important to acknowledge its limitations: 

\begin{itemize}[leftmargin=*]
\item \textit{Task granularity.} The distribution vectors of \emph{DEM} are applicable to data distributions that span a single or multiple tasks. Our experimentation focused on existing data sources with different granularities ranging from several hundred tasks (e.g. P3) to a single one (e.g. MathQA), hence, the resulting distribution vectors captured varying task granularities. A detailed investigation of granularities and how to automatically group the data is an open area of investigation.

\item \textit{Architecture type.} The proposed approach makes no specific assumptions regarding the architecture and should be, in principle, applicable to any architecture variant including Mixture-of-Expert models \cite{fedus2022switch,xue2024openmoe,jiang2024mixtral,sukhbaatar2024branch,hu2024separate}. Due to budget constraints, the evaluation of different architecture types was not included in the experiment plan. Therefore, the compatibility of \emph{DEM} with different architecture types remains to be evaluated. 

\item \textit{Storage Requirements.} \emph{DEM} reduces the computational cost of training models, but it requires storing a number of distribution vectors in the hard drive. For very large models, this creates the need for large storage capacity that may not be always available. One straight-forward solution to this problem is to use parameter-efficient methods to train the distribution vectors instead of full training or discard the distribution vectors once the optimal combination has been identified. 

\end{itemize}

\bibliography{custom}
\bibliographystyle{acl_natbib}

\appendix

\clearpage
\section*{Appendix}
\section{Training Hyperparameters} \label{sec:apx-hyp}
In this section, we describe the detailed hyperparameters that we used for fine tuning the OpenLLaMA model using different datasets separately for \emph{DEM} and combined for \emph{Data Mixing} and \emph{Concatenated Datasets}. In all these cases, we use a constant learning rate of 2e-5 with a 2000 step warmup. We tested other learning rate schedules with cosine and linear decay in preliminary experiments, however, they lead to worse performance. We use AdamW optimizer with $\beta_1=0.9$, $\beta_2=0.95$, weight decay of 0.05 and gradient clipping of 1. We also adjust batch size for different datasets based on the validation loss. We use example packing to fit multiple training examples into a single sample of a batch for efficient training. This is a greedy packing approach where we pack training examples until we reach the max sequence length that we can fit into the model. We do not overflow an example into the next sample of a batch (as generally done during pretraining~\cite{brown2020language}), rather use padding to fill the sample. The full setting is presented in Table~\ref{tab:hyperparameter}. We show the batch size using total number of token after sample packing. The number of training steps indicates the step with best validation loss, and its variable for different datasets. Note that for InstDial, this value is particularly high because of the different kind of samples (i.e. dialog) that consists of this dataset.    

\begin{table}[h]
\centering
\small
\setlength{\tabcolsep}{3pt}
\begin{tabular}{l|ccc} 
\toprule
{\bf Dataset /} & {\bf Batch} & {\bf Learning} & {\bf \# Training}  \\ 
{\bf Method} & {\bf Size} & {\bf Rate} & {\bf Steps}  \\ 
\midrule 
CoT & 65k & 2e-5 & 550 \\
MathQA & 65k & 2e-5 & 600 \\
P3 & 262k & 2e-5 & 6k \\
InstDial & 131k & 2e-5 & 23k \\
SNI & 131k & 2e-5 & 6k \\
Data Mixing & 131k & 2e-5 & 15k \\
Concatenated Datasets & \multirow{1}{*}{131K} & \multirow{1}{*}{2e-5} & \multirow{1}{*}{17K} \\

\bottomrule
\end{tabular}
\caption{Training hyperparameters for different models.}
\label{tab:hyperparameter}
\end{table}

\section{Choosing Data Mixing Weights} \label{sec:apx-data-mix-weights}
Based on initial experiments, we determined the following hyperparameter ranges for the baseline data mixing approach  -- CoT: [0.05, 0.1, 0.15, 0.20], Math QA: [0.05, 0.1, 0.15, 0.20], P3: [0.25, 0.30, 0.35, 0.40], InstructDial: [0.30, 0.35, 0.40, 0.45], Super Natural Instructions: [0.15, 0.20, 0.25, 0.30]. Out of the 1024 possible weight combinations above, we randomly selected 50 combinations for training and selected the best weight setting based on validation-loss. The optimal data mixing setting was the following: P3 - 0.30, SNI - 0.20, Instruct-Dial - 0.40, MathQA - 0.05, CoT - 0.05 The total cost for this hyperparameter search procedure is listed in Table~\ref{tab:train-cost}).

\section{Choosing DEM Weights}
\label{sec:dem-weight}

\begin{table}[t]
\centering
\resizebox{0.8\columnwidth}{!}{%
\setlength{\tabcolsep}{3pt}
\begin{tabular}{l|ccc} 
\toprule
{\bf Models} & {\bf MMLU} & {\bf BBH} & {\bf DROP}  \\ 
\midrule 
Open LLaMA & 40.31 & 32.84 & 24.38 \\
\hline
\multicolumn{4}{c}{{\it DEM} - Distribution Vector} \\
\hline
$\omega = 0.25$ & {50.74} & {\bf 40.56} & {37.96} \\
Random Search, x50 & {\bf 50.98} & 40.55 & {\bf 40.83} \\
\bottomrule
\end{tabular}
}
\caption{Downstream task performance of the DEM w/ Distribution Vector. We compare the weight selection strategies: single-coefficient vs.~random search with 50 iterations.}
\label{tab:random:performance}
\end{table}

In order to select the optimal mixing weights for {\it DEM - Distribution Vector} (Eq~\ref{eq:mdm}), we perform a grid search over $\omega_i$ values. For each coefficient combination we evaluate the validation losses on the five datasets used for fine-tuning (Section~\ref{sec:dataset}), and select the model that minimizes their average. However, exhaustive grid search is expensive as the number of combinations grows exponentially. Thus, we simplify Eq~\ref{eq:mdm} and optimize a single coefficient $\omega$ for all datasets. We found $\omega = 0.25$ (out of 10 values) to produce the best validation loss and use it for all our experiments. 

We chose the weights for {\it DEM - Interpolation} (Eq~\ref{eq:mi}) in a similar manner as {\it DEM - Distribution Vector} by randomly sampling weights from the search grid and normalizing them to sum to 1. Additionally, we also tried the same weights as data mixing and equal weight of 0.2 for each of 5 datasets. The simplest strategy of equal weight performed on par with the best weight combination in terms of average val loss. So, we chose this and reported the corresponding results in Table~\ref{tab:performance}

To measure the effect of using a single coefficient, we perform a limited budget experiment with 50 weight combinations, which are produced using individual weights for each distribution vector (Eq~\ref{eq:ddt}), sampled uniformly from the interval $[0,1]$. Our results show that the best single-coefficient models perform better or on par with the sampled models in terms of average validation loss. This formulation was also adopted in other model interpolation works~\cite{ilharco2022editing, yadav2023tiesmerging}. In Table~\ref{tab:random:performance}, we show the differences in performance on three benchmarks (MMLU, BBH, DROP) using the Open LLaMa 7B model. The two strategies have similar performance on MMLU and BBH but the random search has an advantage of 3 points on DROP. However, this increase comes at the expense of 5x increase in cost (10 evaluations for uniform vs. 50 evaluations for random search). The best distribution weights we found are: CoT - 0.1,  InstDial - 0.12, MathQA - 0.1, P3 - 0.23, SNI - 0.45. We hypothesize that the single-vector weights will not be an optimal choice if there is high negative correlation between the vectors, i.e.,~the data distributions are conflicting.

\section{Fine-Grained Results} \label{sec:apx-perf}
In Table~\ref{tab:perf-mmlu} we show the model performance per domain on the MMLU benchmarking datasets. It covers five different categories, on all of which DEM outperforms the other alternatives.

In Table~\ref{tab:helm_detailed} we show the per-dataset results on HELM benchmark. We can see that our approach significantly outperforms data mixing and improves over the baseline model in most of the categories. Due to space limitations we show different datasets on different rows.

\begin{table*}[t!]
\centering
\resizebox{0.8\textwidth}{!}{%
\setlength{\tabcolsep}{3pt}
\begin{tabular}{l|ccccc} 
\toprule
{\bf Training Dataset} & {\bf STEM} & {\bf Humanities} & {\bf Social Sciences} & {\bf Others} & {\bf Average}  \\ 
\midrule 
Open LLaMA v2 & 33.4 & 36.8 & 45.1 & 47.3 & 40.3 \\
LLaMA 1 \cite{touvron2023llama} & 34.0 & 30.5 & 38.3 & 38.1 & 35.1 \\
LLaMA 2 \cite{touvron2023llama2} & 42.9 & 36.4 & 51.2 & 52.2 & 45.3 \\
\midrule
Public Pool of Prompts (P3) & 25.4 & 32.9 & 44.2 & 41.2 & 35.7  \\
Instruct Dial & 31.9 & 37.8 & 44.5 & 43.5 & 39.3 \\
Super Natural Instructions (SNI) & 38.4 & 42.6 & 53.9 & 52.9 & 46.5 \\
Chain of Thoughts (CoT) & 34.4 & 38.3 & 47.1 & 48.1 & 41.7 \\
Math QA & 32.8 & 36.6 & 44.1 & 46.5 & 39.7 \\
\midrule
Data Mixing & 39.2 & 44.8 & 55.6 & 52.6 & 47.8 \\
Concatenated Datasets (1-5) & 37.9 & 41.4 & 49.8 & 46.1 & 43.4 \\
{\it DEM} - Interpolation (Ours) & 39.7 & 47.2 & 58.5 & 56.2 & 50.1 \\
{\it DEM} - Distribution Vector (Ours) & {\bf 40.4} & {\bf 47.8} & {\bf 58.8} & {\bf 57.0} & {\bf 50.7} \\
\bottomrule
\end{tabular}
}
\caption{MMLU domain specific task performance of models trained on different instruction following datasets (Section~\ref{sec:dataset}). We compare it with different pretrained and fine-tuned baselines (Section~\ref{sec:baseline}) and our proposed approach in Section~\ref{sec:proposal}.}
\label{tab:perf-mmlu}
\end{table*}
\begin{table*}[!t]
\centering

\scriptsize
\resizebox{\textwidth}{!}{%
\setlength{\tabcolsep}{2.5pt}
\begin{tabular}{lcccccccccccc} 
\toprule
\multicolumn{1}{l}{\textbf{Models}} & \multicolumn{1}{c}{{MMLU}} & \multicolumn{1}{c}{{BoolQ}} & \multicolumn{1}{c}{NarrativeQA} & \multicolumn{1}{c}{{NaturalQ closed}} & \multicolumn{1}{c}{NaturalQ open}  & \multicolumn{1}{c}{QUAC} & \multicolumn{1}{c}{TruthfulQA} & \multicolumn{1}{c}{IMDB} & \multicolumn{1}{c}{CivilComments} & \multicolumn{1}{c}{RAFT} & \multicolumn{1}{c}{Wikifact} \\
\midrule 
OpenLLaMA-v2 & 39.37 & 72.3 & 63.96 & 26.08 & 61.15 & 33.03 & 18.65 & 93.2 & 53.96 & 60.0 & 24.89 \\ 
Data Mixing & 43.96 & 85.3 & 71.24 & 21.84 & 19.5 & 34.52 & 42.35 & 87.0 & 64.8 & 69.09 & 21.94 \\ 
DEM (ours) & 46.61 & 82.4 & 71.24 & 28.1 & 69.39 & 40.22 & 29.82 & 96.6 & 53.75 & 67.27 & 26.82 \\ 
\midrule
\multicolumn{1}{l}{\textbf{Models}} & \multicolumn{1}{c}{{ReasonAbstract}} & \multicolumn{1}{c}{{ReasonNatural}} & \multicolumn{1}{c}{bABI} & \multicolumn{1}{c}{{Dyck}} & \multicolumn{1}{c}{GSM-8K}  & \multicolumn{1}{c}{Math-Eq} & \multicolumn{1}{c}{Math-CoT} & \multicolumn{1}{c}{LSAT} & \multicolumn{1}{c}{Legal} & \multicolumn{1}{c}{Imputation} & \multicolumn{1}{c}{EntityMatch}  \\ 
\midrule 

OpenLLaMA-v2 & 18.51 & 21.1 & 45.25 & 52.0 & 5.5 & 12.58 & 8.33 & 20.43 & 48.67 & 81.66 & 83.89 \\
Data Mixing & 20.58 & 29.64 & 54.52 & 40.0 & 0.5 & 8.79 & 1.52 & 24.35 & 62.37 & 76.4 & 85.62 \\ 
DEM (ours) & 23.24 & 34.73 & 56.62 & 48.4 & 6.3 & 11.06 & 4.52 & 18.26 & 58.49 & 71.56 & 85.32 \\ 
\midrule 

\end{tabular}
}

\caption{Detailed HELM results on 22 scenarios. HELM datasets that are part of model-training (BoolQ, GSM-8K and IMDB), are excluded from the aggregated results presented in Table~\ref{tab:helm}} 
\label{tab:helm_detailed}

\end{table*}

\end{document}